\documentclass{article} 
\usepackage{nips12submit_e,times}
\usepackage{hyperref}
\usepackage{graphicx} 
\usepackage{subfigure} 
\usepackage{tabularx}
\usepackage{natbib}
\usepackage{verbatim}
\usepackage{wrapfig}

\usepackage{my_gp_macros}
\usepackage{amsmath}
\usepackage{amssymb}
\usepackage{amsfonts}
\usepackage{parskip}
\usepackage{paralist}

\title{Generalized Product of Experts for Automatic and Principled Fusion of Gaussian Process Predictions}

\author{
	Yanshuai Cao ~~
	David J.~Fleet \\
	Department of Computer Science, ~~
	University of Toronto
}

\newcommand{\Tau}{\mathrm{T}}

\nipsfinalcopy 

\begin{document}
\maketitle

\vspace*{-0.3cm}
\begin{abstract}
\vspace*{-0.2cm}
In this work, we propose a generalized product of
experts (gPoE) framework for combining the predictions of multiple
probabilistic models. We identify four desirable properties that are important for
scalability, expressiveness and robustness, when learning and
inferring with a combination of multiple models. Through analysis and experiments,
we show that gPoE of Gaussian processes (GP) have these qualities, while no other existing combination schemes satisfy
all of them at the same time. The resulting GP-gPoE is highly scalable as individual GP experts can be
independently learned in parallel; very expressive as the way 
experts are combined depends on the input rather than fixed;
the combined prediction is still a valid probabilistic model with
natural interpretation; and finally robust to unreliable predictions
from individual experts.

\end{abstract}

\vspace*{-0.5cm}
\section{Introduction}
\vspace*{-0.2cm}
For both practical and theoretical reasons, it is often necessary to
combine the predictions of multiple learned models. Mixture of
experts, product of experts (PoE)[\cite{hinton2002}], ensemble methods
are perhaps the most obvious frameworks for such prediction fusion.
However there are four desirable properties which no existing fusion
scheme achieves at the same time: \begin{inparaenum}[(i)]\item predictions are combined without the need for joint training or training meta models; \item the way predictions are combined depends on the input rather than fixed; \item the combined prediction is a valid probabilistic model; \item unreliable predictions are automatically filtered out from the combined model.
\end{inparaenum}
Property (i) allows individual experts to be trained independently,
making the overall model easily scalable via parallelization; property
(ii) gives the combined model more expressive power; while property
(iii) allows uncertainty to be used in subsequent modelling or
decision making; and finally property (iv) ensures that the combined
prediction is robust to poor prediction by some of the experts.
In this work, we propose a novel scheme called generalized product of expert (gPoE) that achieves all four properties if individual experts are Gaussian processes, and consequently, excels in terms of scalability, robustness and expressiveness of the resulting model.

In comparison, a mixture of experts with fixed mixing probabilities does not satisfy (ii) and (iv), and because experts and mixing probabilities generally need to be learned together, (i) is not satisfied either. If an input dependent gating function is used, then the MoE can achieve property (ii) and (iv), but joint training is still needed, and the ability (iv) to filter out poor predictions crucially depends on the joint training.

Depending on the nature of the expert model, a PoE may or may not need
joint or re-training, but it does not satisfy property (iv), because
without the gating function to "shut-down" bad experts, the combined
prediction is easily mislead by a single expert putting low probability on the true label.

In the ensemble method regime, bagging\cite{bagging} does not satisfy (ii) and (iv), as it uses fixed equal weights to combine models, and does not automatically filter poor predictions, although empirically it is usually robust due to the equal weight voting. Boosting and stacking\cite{stacking} requires sequential joint training and training a meta-predictor respectively, so they do not satisfy (i). Furthermore, boosting does not satisfy (ii) and (iv), while stacking only has limited ability for (iv) that depends on training. 

As we will demonstrate, the proposed gPoE of Gaussian processes not only enjoys the good
qualities given by the four desired properties of prediction fusion,
but it also retains some important attributes of PoE: many weak
uncertain predictions can yield very sharp combined prediction
together; and the combination has closed analytical form as another
Gaussian distribution.

\vspace*{-0.2cm}
\section{Generalized Product of Expert}
\vspace*{-0.2cm}
\subsection{PoE}
\vspace*{-0.2cm}
We start by briefly describing the product of expert model, of which our proposed method is a generalization. A PoE models a target probability distribution as the product of multiple densities, each of which is given by one expert. The product is then renormalized to sum to one. In the context of supervised learning, the distributions are conditional:
	\begin{equation}
	P(y|x) = \frac{1}{Z}\prod_{i}p_i(y|x)
	\end{equation}

In contrast to mixture models, experts in PoE hold "veto" power, in the sense that a value has low probability under the PoE if a single expert $p_i(y|x)$ assigns low probability to a particular value. As Hinton pointed out \cite{hinton2002}, training such model for general experts by maximizing likelihood is hard because of the renormalization term $Z$. However, in the special case of Gaussian experts $p_i(y|x) = N(m_i(x), \Sigma_i(x))$, the product distribution is still Gaussian, with mean and covariance:
\begin{align}
	m(x) &= (\sum_i m_i(x) \Tau_i(x) ){(\sum_i \Tau_i(x))}^{-1} \label{poe_mean}\\
	\Sigma(x) &= {(\sum_i \Tau_i(x))}^{-1} \label{poe_var}
	\end{align}
where $\Tau_i(x) = \Sigma_i^{-1}(x)$ is the precision of the $i$-th Gaussian expert at point $x$.
Qualitatively, confident predictions have more influence over the
combined prediction than the less confident ones. If the predicted variance were always the correct confidence to be used, then PoE would have exactly the behavior needed. However, a slight model misspecification could cause an expert to produce erroneously low predicted variance along with a biased mean prediction. Because of the combination rule, such over-confidence by a single expert about its erroneous prediction is enough to be detrimental for the resulting combined model.
\vspace*{-0.2cm}
\subsection{gPoE}
\vspace*{-0.2cm}
Given that PoE has almost the desired behavior except for the fact
that an expert's predictive precision is not necessarily the right
measure of reliability of prediction for use in weighting, we will
introduce another measure of such reliability to down-weight or ignore
bad predictions.Like PoE, the proposed generalized product of experts is also a probability model defined by products of distributions. Here we again focus on conditional distribution for supervised learning, taking the form:
	\begin{equation}
	P(y|x) = \frac{1}{Z}\prod_{i}p_i^{\alpha_i(x)}(y|x) \label{gpoe}
	\end{equation}
where $\alpha_i(x) \in \mathbb{R}^{+}$ is a measure of the $i$-th expert's reliability
at point $x$. We will introduce one particular choice for Gaussian
processes in the next subsection, for now let us first analyze the effect of
$\alpha_i(x)$. Raising a density to a power as done in equation
(\ref{gpoe}) has been widely used as a way for annealing
distributions in MCMC, or balancing different parts of a probabilistic model
that has different degrees of freedom (\cite{urtasun_thesis}, see 6.1.2
Balanced GPDM). If $\alpha_i(x) = 1 ~\forall i, x$, then we recover the
PoE as a special case. $\alpha_i > 1$ sharpens the $i$-th distribution
in the product (\ref{gpoe}), whereas $\alpha_i < 1$ broadens it. The
limit cases of $\alpha_i \rightarrow \infty$, while the other
exponents are fixed causes the largest mode of $i$-th distribution to
dominate the product distribution with arbitrarily large ``veto"
power; on the other hand, $\alpha_i \rightarrow 0$ causes $i$-th
expert to have arbitrarily small weight in the combined model,
effectively ignoring its prediction. 

Another interesting property of gPoE is that if each $p_i$ is Gaussian, then the resulting $P(y|x)$ is still Gaussian as in PoE.
To see this, it suffices to show that $p_i^{\alpha_i}$ is Gaussian, which is apparent with a little algebraic manipulation:
\begin{equation}
p_i^{\alpha_i} = \exp({\alpha_i}\ln(p_i)) = \frac{1}{C}\exp(-.5(y-m_i)^{\T}(\alpha_i\Sigma_i^{-1})(y-m_i))
\end{equation}
This also shows that the power $\alpha_i$ essentially scales the precision of $i$-th Gaussian. Therefore, similar to equation (\ref{poe_mean}) and (\ref{poe_var}), the mean and covariance of Gaussian gPoE are:
\begin{align}
	m(x) &= (\sum_i m_i(x) \alpha_i(x)\Tau_i(x) ){(\sum_i \alpha_i(x)\Tau_i(x))}^{-1} \label{gpoe_mean}\\
	\Sigma(x) &= {(\sum_i \alpha_i(x)\Tau_i(x))}^{-1} \label{gpoe_var}
	\end{align}

\vspace*{-0.2cm}
\subsection{gPoE for Gaussian processes}
\vspace*{-0.2cm}
Now that we have established how $\alpha_i(x)$ can be used to control the influence
of individual experts, there is a natural choice of $\alpha_i(x)$ for Gaussian processes that can reliably detect  
if a particular GP expert does not generalize well at a given point
$x$: the change in entropy from prior to posterior at point $x$,
$\Delta H_i(x)$, which takes almost no extra computation since
posterior variance at $x$ is already computed when the GP expert makes
prediction and the prior variance is simply $k(x,x)$, where $k$ is the
kernel used.

When the entropy change at point $x$ is zero, it means the $i$-th
expert does not have any information about this point that comes from
training observation, therefore, it shall not contribute to the
combined prediction, which is achieved by our model because
$\alpha_i(x) = 0$ in (\ref{gpoe_mean}) and (\ref{gpoe_var}). For
Gaussian processes, this covers both the case if point $x$ is far away
from training points or if the model is misspecified. \footnote{For the experiments, we also normalize the weighting factors at each point $x$ so that $\sum_i{\alpha_i(x)=1}$.}

There are other quantities that could be used as $\alpha_i(x)$, for
example the difference between the prior and posterior variance
(instead of half of difference of log of the two variances). The
reason we choose the entropy change is because it is unit-less in the
sense of dimensional analysis in physics, so that the resulting
predictive variance in equation (\ref{gpoe_var}) is of the correct
unit. The same is not true if $\alpha_i(x)$ is the difference of
variances which carries the unit of variance (or squared unit of
variable $y$). The KL divergence between prior and posterior
distribution at point $x$ could also potentially be used as
$\alpha_i(x)$, but we find the entropy change to be already effective
in our experiments. We will explore the use of KL divergence in future
work.

\vspace*{-0.2cm}
\section{Experiment}
\vspace*{-0.2cm}

We compare gPoE against bagging, MoE, and PoE on three different
datasets: KIN40K (8D feature space, 10K training points), SARCOS
(21D, 44484 training points), and the UK apartment price dataset (2D, 64910
training points) used in SVI-GP
work of Hensman et al.\ \cite{hensman}. We try three different
ways to build individual GP experts: ({\em SoD}) random subset of data;
({\em local}) local GP around a randomly selected point; ({\em tree}) a tree based
construction, where a ball tree \cite{balltree} built on training set
recursively partitions the space, and on each level of the tree, a
random subset of data is drawn to build a GP. On all datasets and for
all methods of GP expert construction, we use $256$ data points for
each expert, and construct $512$ GP experts in total. Each GP expert
uses a kernel that is the sum of an ARD kernel and white kernel, and
all hyperparameters are learned by scaled conjugate gradient.

For MoE, we do not jointly learn experts and gating functions, as it
is very time consuming, instead we use the same entropy change as the
gating function (re-normalized to sum to one). Therefore, all experts
in all combination schemes could be learned independently in parallel.
On a 32-core machine, with the described setup, training $512$ GP
experts with independent hyperparameter learning via SGD on each
datasets takes between 20 seconds to just under one minute, including
the time for any preprocessing such as fitting the ball tree.

In terms of test performance evaluation, we use the commonly used
metrics standardized negative log probability (SNLP) and standardized
mean square error (SMSE). Table \ref{smse} and \ref{snlp} show that
gPoE consistently out-performs bagging, MoE and PoE combination rules
by a large margin on both scores. Under the tree based expert
construction method, we explore a heuristic variant of gPoE
(tree-gPoE), which takes only experts that are on the path defined by the root node to
the leaf node of the test point in the tree. Alternatively, this could
be viewed as defining $\alpha_i(x)=0$ for all experts that are not on
the root-to-leaf path. This variant gives a slight further boost to
performance across the board.

Another interesting observation is that while gPoE performs consistently
well, PoE is almost always poor, especially in SNLP score. This
empirically confirms our previous analysis that misguided
over-confidence by experts are detrimental to the resulting PoE, and
shows that the correction by entropy change in gPoE is an effective
way to fix this problem.

\vspace*{-0.2cm}
\begin{small}
\begin{table}[h]
\centering
\begin{tabular}{cccccccccccccc}
\hline
 & \multicolumn{4}{c||}{\tiny{SoD}} &
\multicolumn{4}{c||}{\tiny{Local}} & \multicolumn{5}{c}{\tiny{Tree}}
\\ \hline 

\multicolumn{1}{|c|}{\tiny{}} & 
\multicolumn{1}{c|}{\!\!\!\tiny{Bagging}\!\!\!} &
\multicolumn{1}{c|}{\!\!\!\tiny{MoE}\!\!\!} & 
\multicolumn{1}{c|}{\!\!\!\tiny{PoE}\!\!\!} &
\multicolumn{1}{c||}{\!\!\!\tiny{gPoE}\!\!\!} & 
\multicolumn{1}{c|}{\!\!\!\tiny{Bagging}\!\!\!} &
\multicolumn{1}{c|}{\!\!\!\tiny{MoE}\!\!\!} & 
\multicolumn{1}{c|}{\!\!\!\tiny{PoE}\!\!\!} &
\multicolumn{1}{c||}{\!\!\!\tiny{gPoE}\!\!\!} & 
\multicolumn{1}{c|}{\!\!\!\tiny{Bagging}\!\!\!}
& \multicolumn{1}{c|}{\!\!\!\tiny{MoE}\!\!\!} & 
\multicolumn{1}{c|}{\!\!\!\tiny{PoE}\!\!\!} &
\multicolumn{1}{c|}{\!\!\!\tiny{gPoE}\!\!\!} &
\multicolumn{1}{c|}{\!\!\!\tiny{tree-gPoE}\!\!\!} \\ \hline

\multicolumn{1}{|c|}{\tiny{SARCOS}} & 
\multicolumn{1}{c|}{\tiny{.619}} &
\multicolumn{1}{c|}{\tiny{0.164}} & 
\multicolumn{1}{c|}{\tiny{ 0.438}} & 
\multicolumn{1}{c||}{\tiny{\textbf{0.0603}}} &
\multicolumn{1}{c|}{\tiny{0.685}} & 
\multicolumn{1}{c|}{\tiny{0.119}} & 
\multicolumn{1}{c|}{\tiny{0.619}} &
\multicolumn{1}{c||}{\tiny{\textbf{0.0549}}} &
\multicolumn{1}{c|}{\tiny{0.648}} &
\multicolumn{1}{c|}{\tiny{0.208}} & 
\multicolumn{1}{c|}{\tiny{0.493}} &
\multicolumn{1}{c|}{\tiny{0.014}} & 
\multicolumn{1}{c|}{\tiny{\textbf{0.009}}} \\ \hline

\multicolumn{1}{|c|}{\tiny{KIN40K}} & 
\multicolumn{1}{c|}{\tiny{0.628}} &
\multicolumn{1}{c|}{\tiny{0.520}} & 
\multicolumn{1}{c|}{\tiny{0.543}} & 
\multicolumn{1}{c||}{\tiny{\textbf{0.346}}}&
\multicolumn{1}{c|}{\tiny{0.761}} & 
\multicolumn{1}{c|}{\tiny{1.174}} &
\multicolumn{1}{c|}{\tiny{0.671}} & 
\multicolumn{1}{c||}{\tiny{\textbf{0.381}}}& 
\multicolumn{1}{c|}{\tiny{0.735}} &
\multicolumn{1}{c|}{\tiny{0.691}} & 
\multicolumn{1}{c|}{\tiny{0.652}} &
\multicolumn{1}{c|}{\tiny{0.285}} & 
\multicolumn{1}{c|}{\tiny{\textbf{0.195}}} \\ \hline 

\multicolumn{1}{|c|}{\tiny{UK-APT}\!\!\!} & 
\multicolumn{1}{c|}{\!\!\!\tiny{0.00219}\!\!\!} &
\multicolumn{1}{c|}{\!\!\!\tiny{0.00220}\!\!\!} & 
\multicolumn{1}{c|}{\!\!\!\tiny{0.00218}\!\!\!} & 
\multicolumn{1}{c||}{\!\!\!\tiny{\textbf{0.00214}}\!\!\!}&
\multicolumn{1}{c|}{\!\!\!\tiny{0.00316}\!\!\!} & 
\multicolumn{1}{c|}{\!\!\!\tiny{0.00301}\!\!\!} &
\multicolumn{1}{c|}{\!\!\!\tiny{0.00315}\!\!\!} & 
\multicolumn{1}{c||}{\!\!\!\tiny{\textbf{0.00122}}\!\!\!}& 
\multicolumn{1}{c|}{\!\!\!\tiny{0.00309}\!\!\!} &
\multicolumn{1}{c|}{\!\!\!\tiny{0.00193}\!\!\!} & 
\multicolumn{1}{c|}{\!\!\!\tiny{0.00310}\!\!\!} &
\multicolumn{1}{c|}{\!\!\!\tiny{0.00162}\!\!\!} & 
\multicolumn{1}{c|}{\!\!\!\tiny{\textbf{0.00144}}\!\!\!} \\ \hline 
\end{tabular}
\caption{\small{SMSE}}\label{smse}
\end{table}
\end{small}
\vspace*{-0.2cm}
\begin{table}[h]
\centering
\begin{tabular}{cccccccccccccc}
\hline
 & \multicolumn{4}{c||}{\tiny{SoD}} &
\multicolumn{4}{c||}{\tiny{Local}} & \multicolumn{5}{c}{\tiny{Tree}}
\\ \hline 

\multicolumn{1}{|c|}{\tiny{}} & 
\multicolumn{1}{c|}{\!\!\!\tiny{Bagging}\!\!\!} &
\multicolumn{1}{c|}{\!\!\!\tiny{MoE}\!\!\!} & 
\multicolumn{1}{c|}{\!\!\!\tiny{PoE}\!\!\!} &
\multicolumn{1}{c||}{\!\!\!\tiny{gPoE}\!\!\!} & 
\multicolumn{1}{c|}{\!\!\!\tiny{Bagging}\!\!\!} &
\multicolumn{1}{c|}{\!\!\!\tiny{MoE}\!\!\!} & 
\multicolumn{1}{c|}{\!\!\!\tiny{PoE}\!\!\!} &
\multicolumn{1}{c||}{\!\!\!\tiny{gPoE}\!\!\!} & 
\multicolumn{1}{c|}{\!\!\!\tiny{Bagging}\!\!\!}
& \multicolumn{1}{c|}{\!\!\!\tiny{MoE}\!\!\!} & 
\multicolumn{1}{c|}{\!\!\!\tiny{PoE}\!\!\!} &
\multicolumn{1}{c|}{\!\!\!\tiny{gPoE}\!\!\!} &
\multicolumn{1}{c|}{\!\!\!\tiny{tree-gPoE}\!\!\!} \\ \hline

\multicolumn{1}{|c|}{\tiny{SARCOS}} & 
\multicolumn{1}{c|}{\tiny{N\slash A}\!\!\!} &
\multicolumn{1}{c|}{\tiny{-0.528}} & 
\multicolumn{1}{c|}{\tiny{205.27}} & 
\multicolumn{1}{c||}{\tiny{\textbf{-1.445}}}
& \multicolumn{1}{c|}{\tiny{N\slash A}\!\!\!} &
\multicolumn{1}{c|}{\tiny{-1.432}} & 
\multicolumn{1}{c|}{\tiny{\tiny{3622.5}}} &
\multicolumn{1}{c||}{\tiny{\textbf{-2.456}}} &
\multicolumn{1}{c|}{\tiny{N\slash A}\!\!\!}
& \multicolumn{1}{c|}{\tiny{-0.896}} & 
\multicolumn{1}{c|}{\tiny{1305.46}} &
\multicolumn{1}{c|}{\tiny{-2.643}} & 
\multicolumn{1}{c|}{\tiny{\textbf{-2.77}}} \\ \hline 

\multicolumn{1}{|c|}{\tiny{KIN40K}} & 
\multicolumn{1}{c|}{\tiny{N\slash A}\!\!\!} &
\multicolumn{1}{c|}{\tiny{-0.344}} & 
\multicolumn{1}{c|}{\tiny{215.02}} & 
\multicolumn{1}{c||}{\tiny{\textbf{-0.542}}}&
\multicolumn{1}{c|}{\tiny{N\slash A}\!\!\!} & 
\multicolumn{1}{c|}{\tiny{0.6136}} &
\multicolumn{1}{c|}{\tiny{495.17}} & 
\multicolumn{1}{c||}{\tiny{\textbf{-0.518}}} & 
\multicolumn{1}{c|}{\tiny{N\slash A}\!\!\!} &
\multicolumn{1}{c|}{\tiny{-0.155}} & 
\multicolumn{1}{c|}{\tiny{376.4}} &
\multicolumn{1}{c|}{\tiny{-0.643}} & 
\multicolumn{1}{c|}{\tiny{\textbf{-0.824}}} \\ \hline 

\multicolumn{1}{|c|}{\tiny{UK-APT}} & 
\multicolumn{1}{c|}{\tiny{N\slash A}\!\!\!} &
\multicolumn{1}{c|}{\tiny{-0.175}} & 
\multicolumn{1}{c|}{\tiny{244.06}} & 
\multicolumn{1}{c||}{\tiny{\textbf{-0.191}}}
& \multicolumn{1}{c|}{\tiny{N\slash A}\!\!\!} & 
\multicolumn{1}{c|}{\tiny{-0.215}} &
\multicolumn{1}{c|}{\tiny{805.4}} & 
\multicolumn{1}{c||}{\tiny{\textbf{-0.337}}} & 
\multicolumn{1}{c|}{\tiny{N\slash A}\!\!\!}
& \multicolumn{1}{c|}{\tiny{-0.235}} & 
\multicolumn{1}{c|}{\tiny{627.07}} &
\multicolumn{1}{c|}{\tiny{-0.355}} & 
\multicolumn{1}{c|}{\tiny{\textbf{-0.410}}} \\ \hline 
\end{tabular}
\caption{\small{SNLP}}\label{snlp}
\end{table}

Finally, we would like to note that as a testimony to the expressive
power given by the gPoE, GP experts trained on only $256$ points with very
generic kernels could combine to give prediction performance close to or even superior to
sophisticated sparse Gaussian process approximation such as stochastic
variational inference (SVI-GP), as evidenced by the comparison in
table \ref{rmse} for the UK-APT dataset. Note also that due to
parallelization, training in our case took less than 30 seconds on
this problem of 64910 training points, although testing time in our gPoE
is much longer than sparse GP approximations. Similar results
competitive or even superior to sophisticated FITC\cite{Snelson06} or CholQR\cite{cholqr}
approximations are observed on SARCOS and KIN40K as well, but due to
space and time constraints are not included in this extended
abstract, but in future work instead. We would like to emphasize that
such comparison does not suggest gPoE as a silver bullet for beating
benchmarks using any naive expert GP model, but to demonstrate the expressiveness of the resulting model, and shows
its potential to be used in conjunction with other sophisticated
techniques for sparsification and automatic model selection.

\begin{table}[h]
\centering
\begin{tiny}
\begin{tabular}{|c|c|c|c|c|c|c|c|c|c|c|}
\hline
 & \!\!\!SoD-256\!\!\! & \!\!\!SoD-500$^{*}$\!\!\! & \!\!\!SoD-800$^{*}$\!\!\! & \!\!\!SoD-1000$^{*}$\!\!\! & \!\!\!SoD-1200$^{*}$\!\!\! & \!\!\!SVI-GP$^{*}$\!\!\! & \!\!\!SoD-gPoE\!\!\! & \!\!\!Local-gPoE\!\!\! & \!\!\!Tree-gPoE\!\!\! & \!\!\!Tree2-gPoE\!\!\! \\ \hline
RMSE & 0.566 & \!\!\!0.522 +/- 0.018\!\!\! &  \!\!\!0.510 +/- 0.015\!\!\! & \!\!\!0.503 +/- 0.011\!\!\! &
\!\!\!0.502 +/- 1.012\!\!\! & 0.426 & 0.556 & \textbf{0.419} &  0.484 &  0.456 \\ \hline
\end{tabular}

\caption{\small{RMSE: comparing root mean square error on the UK-APT dataset,
  we use this score instead of SMSE and SNLP as it is the measure used by Hensman et al.\ in \cite{hensman}.
All methods with $*$ next to the name indicates that the number is
what is reported in the SVI-GP paper \cite{hensman}. }}\label{rmse}
\end{tiny}
\end{table}

\vspace*{-0.2cm}
\section{Discussion and Conclusion}
\vspace*{-0.2cm}
In this work, we proposed a principled way to combine predictions of
multiple independently learned GP experts without the need for further
training. The combined model takes the form of a generalized product
of experts, and the combined prediction is Gaussian and has desirable
properties such as increased expressiveness and robustness to poor
predictions of some experts. 

We showed that the gPoE has many interesting qualities over other
combination rules. However, one thing it cannot capture is
multi-modality like in mixture of experts. In future work it would
interesting to explore generalized product of mixture of Gaussian
processes, which captures both ``or" constraints as well as ``and"
constraints. Another future work direction is to explore other measures
of model reliability for GP. Finally, while the lack of change in
entropy is an indicator of irrelevant prediction, the converse
statement does not seem to be true, i.e., sufficient change in
entropy does not necessarily guarantee reliable prediction because of
all the potential ways the model could be mis-specified. However, our
empirical results suggest that at least with RBF kernels, the change
in entropy is always reliable even if the estimated posterior variance itself
is not accurate. Further theoretical work is needed to better
understand the converse case.

\vspace*{-0.2cm}
\subsection*{Acknowledgments}
\vspace*{-0.2cm}
We thank J. Hensman for providing his dataset for benchmarking, as
well as the anonymous reviewer for insightful comments and question
about the reliability issue in the converse case.



\end{document}